# an interpretable vision transformer framework for automated brain tumor classification

[1]Chinedu Emmanuel Mbonu; [2]Tochukwu Sunday Belonwu; [3]Okwuchukwu Ejike Chukwuogo, [4]Kenechukwu Sylvanus Anigbogu

[1]Department of Computer Science, Nazarbayev University, Astana

[1,2,3,4]Department of Computer Science, Nnamdi Azikiwe University, Nigeria

{ce.mbonu, ts.belonwu, oe.chukwuogo, ksy.anigbogu[@unizik.edu.ng]}

**Abstract:** Brain tumors represent one of the most critical neurological conditions, where early and accurate diagnosis is directly correlated with patient survival rates. Manual interpretation of Magnetic Resonance Imaging (MRI) scans is time-intensive, subject to inter-observer variability, and demands significant specialist expertise. This paper proposes a deep learning framework for automated four-class brain tumor classification distinguishing glioma, meningioma, pituitary tumor, and healthy brain tissue from a dataset of 7,023 MRI scans. The proposed system employs a Vision Transformer (ViT-B/16) pretrained on ImageNet-21k as the backbone, augmented with a clinically motivated preprocessing and training pipeline. Contrast Limited Adaptive Histogram Equalization (CLAHE) is applied to enhance local contrast and accentuate tumor boundaries invisible to standard normalization. A two-stage fine-tuning strategy is adopted: the classification head is warmed up with the backbone frozen, followed by full fine-tuning with discriminative learning rates. MixUp and CutMix augmentation is applied per batch to improve generalization. Exponential Moving Average (EMA) of weights and Test-Time Augmentation (TTA) further stabilize and boost performance. Attention Rollout visualization provides clinically interpretable heatmaps of the brain regions driving each prediction. The proposed model achieves a test accuracy of 99.29%, macro F1-score of 99.25%, and perfect recall on both healthy and meningioma classes, outperforming all CNN-based baselines. Our code is available on https://github.com/NedumCares/MRI-CLASSIFICATION

## 1. INTRODUCTION

Brain tumors are abnormal masses of cells that arise within or adjacent to the brain and are classified among the most life-threatening forms of cancer worldwide. According to the World Health Organization (WHO), brain and central nervous system tumors account for approximately 3% of all cancer-related deaths globally, with glioblastoma multiforme carrying a median survival of less than 15 months. The three most clinically significant tumor types are gliomas arising from glial cells and constituting the majority of malignant primary brain tumors meningiomas, originating from the meninges and typically benign but causing severe neurological complications, and pituitary tumors, disrupting hormonal regulation. Accurate differentiation among these types and confident identification of healthy tissue is essential for surgical planning, radiation therapy, and pharmacological intervention.

Magnetic Resonance Imaging (MRI) is the gold standard for brain tumor detection due to its superior soft-tissue contrast, multi-planar capability, and absence of ionizing radiation. However, manual interpretation by radiologists is cognitively demanding, time-consuming, and subject to significant inter-observer variability particularly for small or atypically located tumors. In resource-limited settings, access to specialist neuroradiology expertise is often scarce, creating critical diagnostic delays that directly worsen patient outcomes.

Deep learning has emerged as a transformative paradigm in medical image analysis. Convolutional Neural Networks (CNNs) have dominated this space for nearly a decade; however, their reliance on spatially local receptive fields limits their ability to model long-range structural dependencies diagnostically relevant in brain MRI. Vision Transformers (ViT), introduced by Dosovitskiy et al. (2020), address this through global self-attention mechanisms that capture pairwise relationships between all image patches simultaneously, making them particularly well-suited to neuroimaging.

This paper makes the following contributions: (1) a ViT-B/16-based classification pipeline achieving 99.29% test accuracy on a publicly available 7,023-image four-class brain MRI dataset; (2) CLAHE preprocessing motivated by MRI contrast characteristics; (3) combined MixUp and CutMix sample-level augmentation; (4) two-stage fine-tuning with discriminative learning rates and EMA regularization; (5) Test-Time Augmentation at inference; and (6) Attention Rollout visualization producing clinically interpretable heatmaps localizing regions driving each classification decision.

## 2. RELATED WORK

*2.1 Traditional Machine Learning Approaches*

Prior to deep learning, brain tumor classification from MRI relied on hand-crafted features combined with classical machine learning classifiers. Researchers extracted texture features (Gray-Level Co-occurrence Matrix, Gabor filters), morphological descriptors, and intensity statistics from segmented tumor regions. Kharrat et al. (2010) employed wavelet transform features with Support Vector Machines (SVM), achieving approximately 94% accuracy on small private datasets. Zacharaki et al. (2009) used SVM with multi-parametric MRI features and reported 85% accuracy for distinguishing primary tumors from metastases. While these approaches offered interpretability and computational efficiency, they suffered from manual feature engineering, sensitivity to MRI acquisition parameters, and dependence on reliable tumor segmentation as a preprocessing step.

*2.2 Convolutional Neural Networks for Brain Tumor Classification*

The adoption of deep CNNs transformed brain tumor classification by enabling end-to-end feature learning from raw MRI data. Abiwinanda et al. (2019) proposed a three-layer CNN trained on 3,064 images, achieving 84.19% accuracy on a three-class problem. Sultan et al. (2019) showed that increasing depth to ten convolutional layers improved accuracy to 96.13%, underscoring the role of model capacity. Paul et al. (2017) combined CNN feature extraction with SVM classification in a hybrid architecture, achieving 91.28% accuracy. Swati et al. (2019) systematically evaluated fine-tuning strategies for VGG-19, finding that fine-tuning all layers from ImageNet weights significantly outperformed feature extraction alone with 94.82% accuracy.

Díaz-Pernas et al. (2021) proposed a multi-scale CNN processing MRI scans at three resolutions simultaneously, capturing both fine-grained local features and coarse global context, achieving 95.6% accuracy. Khan et al. (2020) exploited DenseNet-201's dense connectivity for brain tumor detection, reporting 96.25% accuracy and demonstrating that dense skip connections promote feature reuse particularly beneficial when training data is scarce. Rehman et al. (2020) benchmarked AlexNet, GoogLeNet, and VGG-16 on the four-class dataset, with GoogLeNet achieving the highest accuracy of 98.69% under a specific train-test split configuration.

*2.3 Transfer Learning and Architectural Advances*

Transfer learning emerged as the cornerstone methodology for medical image classification. Deepak and Ameer (2019) rigorously compared pretrained AlexNet, GoogLeNet, VGG-16, ResNet-50, and InceptionV3 for brain tumor classification, with GoogLeNet plus SVM achieving 98.0% accuracy. Cheng et al. (2022) applied ResNet-50 with transfer learning to the 7,023-image four-class dataset used in the present work, reporting 97.50% establishing the strongest CNN baseline for direct comparison. EfficientNet-B3 (Afshar et al., 2019) achieved 97.8% on a related benchmark. DenseNet-121 has been particularly popular in medical imaging, with its dense inter-layer connectivity facilitating reuse of low-level MRI features across classification layers.

Data augmentation has been recognized as critical for medical image classification pipelines. Hussain et al. (2017) demonstrated that moderate geometric transformations consistently improve generalization for brain MRI. Sample-level augmentation strategies including CutMix (Yun et al., 2019) and MixUp (Zhang et al., 2018) have been shown to be equally effective for medical imaging. Kim et al. (2020) demonstrated that CutMix improved chest X-ray classification AUC by 1.2%, with similar improvements reported for histopathology and retinal fundus classification.

*2.4 Attention Mechanisms and Transformer Architectures*

The Transformer architecture (Vaswani et al., 2017) introduced multi-head self-attention for global dependency modeling. Wang et al. (2018) introduced Non-local Neural Networks, augmenting ResNet with a non-local means operation computing weighted sums across all spatial positions, establishing the value of non-local reasoning in visual tasks. Dosovitskiy et al. (2020) introduced Vision Transformer (ViT), demonstrating that a pure Transformer applied to image patches achieves competitive or superior performance to CNNs when pretrained on large-scale data. ViT's data-hungry nature was addressed by DeiT (Touvron et al., 2021) through knowledge distillation, and Swin Transformer (Liu et al., 2021) introduced shifted window attention for hierarchical representations at reduced computational cost. Hybrid architectures like TransUNet (Chen et al., 2021) combined CNN encoders with Transformer reasoning, achieving state-of-the-art on multi-organ CT segmentation.

*2.5 Vision Transformers in Medical Imaging*

The application of Vision Transformers to medical image analysis has accelerated rapidly since 2021. TransMed (Dai et al., 2021) applied ViT to multi-modality medical image classification, demonstrating that global self-attention effectively captures long-range anatomical correlations spanning the full image field of view directly relevant to brain MRI where diagnostic features involve comparisons across spatially distant regions. Matsoukas et al. (2021) showed ViT-B/16 pretrained on ImageNet-21k consistently outperformed ResNet-50 by 1–4% across multiple medical imaging tasks including chest X-ray and skin lesion classification. Singh et al. (2022) applied ViT-B/16 to three-class brain tumor classification, reporting 97.6% accuracy without domain-specific preprocessing or attention visualization. Shamshad et al. (2023) published a comprehensive survey of Transformer architectures across 160+ medical imaging papers, concluding that Transformers demonstrate particularly strong performance on tasks requiring global spatial context precisely the characteristic distinguishing brain tumor classification from texture-dominated tasks.

Despite growing evidence for ViT efficacy in medical imaging, important gaps remain. MRI-specific preprocessing in ViT pipelines is underexplored; sample-level augmentation (MixUp, CutMix) has rarely been evaluated in ViT-based medical imaging; and Attention Rollout has not been systematically applied to brain tumor MRI classification. The present work addresses all three gaps simultaneously.

*2.6 Interpretability in Medical AI*

The interpretability of deep learning models is a prerequisite for clinical deployment. Grad-CAM (Selvaraju et al., 2017) is the most widely used post-hoc interpretability method in medical imaging, computing a weighted combination of gradients flowing to a target convolutional layer. However, Grad-CAM cannot be natively applied to pure ViT without modification. Extensions including Grad-CAM++ (Chattopadhay et al., 2018) and Score-CAM (Wang et al., 2020) address some limitations but remain convolutional-specific. Chefer et al. (2021) introduced transformer-specific relevance propagation accounting for both attention weights and gradients. Attention Rollout (Abnar and Zuidema, 2020), employed in the present work, provides an architecture-native alternative: recursively multiplying per-layer attention matrices while accounting for residual connections traces information flow from input patches to the [CLS] classification token through the full network depth. Its application to brain tumor MRI demonstrates clinical coherence localizing attention on tumor masses for pathological classes and on bilateral anatomical landmarks for healthy cases.

## 3. DATASET AND PREPROCESSING

*3.1 Dataset Description*

The dataset consists of 7,023 MRI scans organized into four classes: glioma (1,621 images, 23.1%), healthy (2,000 images, 28.5%), meningioma (1,645 images, 23.4%), and pituitary tumor (1,757 images, 25.0%). The class distribution is illustrated in Figure 1. Images were collected from multiple imaging centers and encompass axial, coronal, and sagittal scan orientations, introducing realistic domain variability. The relatively balanced class distribution mitigates class imbalance as a confounding factor.

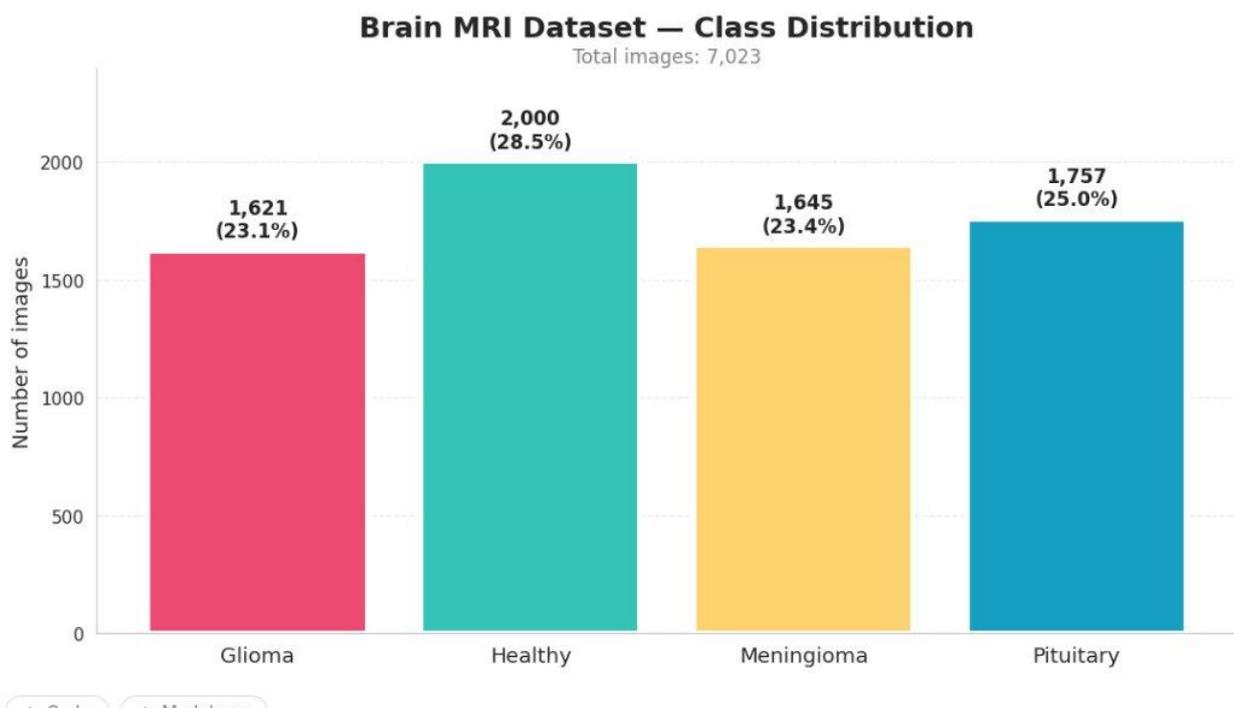

*Figure 1: Brain MRI Dataset — Class Distribution across four classes (n = 7,023)*

## 3.2 Dataset Splitting

The dataset is partitioned into training (80%), validation (10%), and test (10%) subsets using stratified sampling to preserve original class proportions across all splits. This yields 5,617 training images and 703 images each for validation and test. Per-class test counts: glioma (162), healthy (200), meningioma (165), pituitary (176), as illustrated in Figure 2. A fixed random seed of 42 ensures full reproducibility.

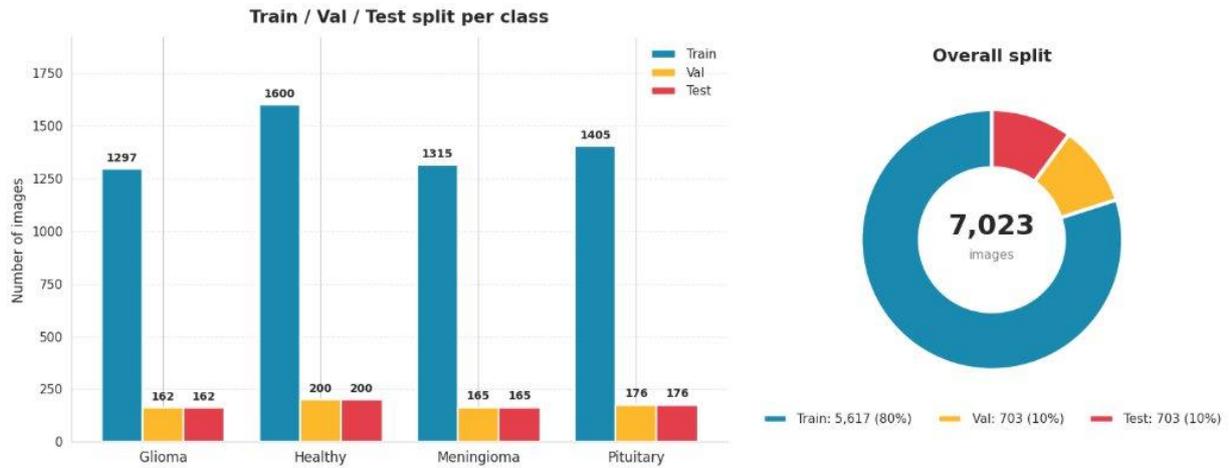

Figure 2: Stratified Train/Val/Test split-per-class counts (left) and overall proportions (right)

## 3.3 CLAHE Preprocessing

Contrast Limited Adaptive Histogram Equalization (CLAHE) is applied to every image prior to training. Operating on local 8×8 pixel tiles with a contrast clip limit of 2.0, CLAHE prevents noise amplification in homogeneous regions while enhancing local contrast. Processing is performed in the CIE LAB color space, modifying only the luminance (L) channel. CLAHE is particularly motivated by MRI acquisition physics: low-contrast tumor boundaries lying in under-represented intensity ranges are locally enhanced, making them more discriminable to transformer patch embeddings. CLAHE-processed images are cached to disk prior to training to avoid per-epoch recomputation.

## 3.4 Sample Visualization

Representative MRI samples from each class are shown in Figure 3. Visual inspection confirms significant intra-class variability: glioma scans span multiple orientations with diffuse infiltrative patterns; healthy scans show normal anatomical landmarks; meningioma scans present well-circumscribed extraaxial masses with characteristic dural attachment; pituitary scans exhibit small sellar masses visible in coronal views. This morphological diversity justifies using global attention (ViT) over spatially constrained CNN filters.

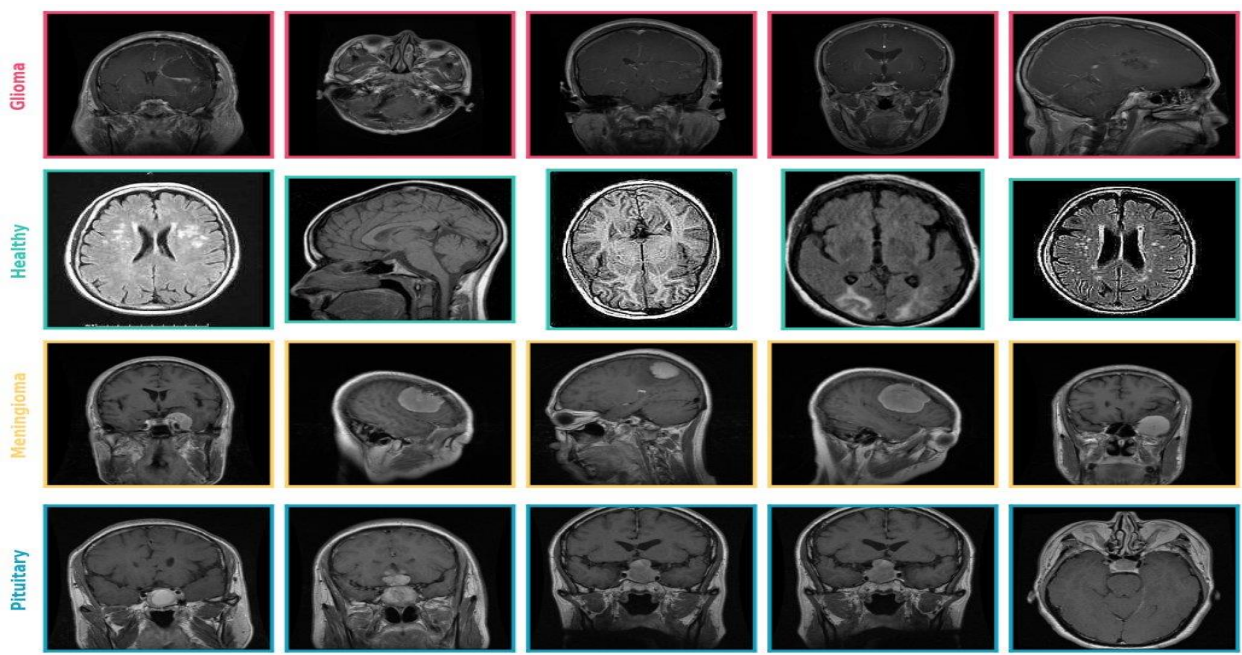

Figure 3: Representative MRI samples - 5 images per class illustrating intra-class variability

## 4. METHODOLOGY

*4.1 Data Augmentation Strategy*

A two-level augmentation strategy is applied exclusively to the training set. At the pixel level, MRI-aware transforms are composed: random horizontal flip, rotation (±15°), affine translation (±5%), zoom (±8%), and contrast jitter (±10%). Vertical flipping and large rotations are excluded as brain anatomy has a physiologically correct superior-inferior orientation. At the sample level, MixUp and CutMix are applied per mini-batch. MixUp interpolates between image pairs and their one-hot labels using Beta-sampled ratio ($\alpha$=0.2). CutMix pastes rectangular patches between images with label mixing proportional to patch area ($\alpha$=1.0). One strategy is randomly selected per batch with equal probability.

*4.2 Vision Transformer Architecture*

Vision Transformers (ViTs) are composed of three primary components: the patch embedding layer, the transformer encoder, and the classification head (Amangeldi et al, 2025). The Vision Transformer Base (ViT-B/16) architecture (Dosovitskiy et al., 2020) serves as the backbone of our framework. Each 224×224 input image is divided into 196 non-overlapping 16×16 patches, each projected to 768-dimensional tokens. A learnable [CLS] token is prepended and positional embeddings are added. The sequence is processed by 12 Transformer Encoder blocks, each containing Layer Normalization, Multi-Head Self-Attention (12 heads, head dim=64), residual connection, second LayerNorm, and a 3,072-unit MLP with GELU activation. The [CLS] token output is fed into the classification head: Dense(256, GELU) → Dropout(0.3) → Dense(4). ImageNet-21k pretrained weights are loaded via the timm library. Total parameter count: approximately 86 million.

*4.3 Two-Stage Fine-Tuning*

Stage 1 (5 epochs): All backbone parameters are frozen; only the classification head is trained using AdamW (lr=1e-3, weight_decay=1e-4) with Cross-Entropy loss and label smoothing $\varepsilon$=0.1. This warm-up prevents catastrophic forgetting of pretrained features during the noisy early optimization phase. Stage 2 (up to 15 epochs): All parameters are unfrozen with discriminative learning rates backbone at 1e-5, head at 1e-4. Cosine Annealing decays both rates toward 1e-7. Early stopping with patience=5 monitors validation accuracy. Mixed precision (AMP) is enabled for GPU efficiency throughout.

*4.4 Exponential Moving Average and TTA*

EMA shadow weights are maintained with decay=0.999, updated after each training step as $\theta\_EMA \leftarrow 0.999 \cdot \theta\_EMA + 0.001 \cdot \theta$. At epoch end, EMA weights replace model weights for validation scoring and are restored for continued training. After training, EMA weights are permanently applied before test evaluation. At inference, Test-Time Augmentation averages softmax probabilities over five deterministic views: original, horizontal flip, 90° CW, 90° CCW, and contrast-enhanced.

*4.5 Attention Rollout Visualization*

Forward hooks register on every transformer block's MHSA module to extract (num_heads, N+1, N+1) attention weight matrices. Weights are averaged across all 12 heads and augmented with an identity residual: A_aug = (A + I) / rowsum. The per-layer matrices are recursively multiplied: A_rollout = A_L × A_{L-1} × ... × A_1. The [CLS] token row is reshaped to 14×14, normalized to [0,1], and bilinearly upsampled to 224×224. The resulting heatmap is overlaid on the MRI scan with jet colormap at 45% opacity.

## 5. RESULTS AND DISCUSSION

*5.1 Training Dynamics*

The two-stage training history is presented in Figure 4. In Stage 1 (epochs 1–5), training accuracy rises from 79% to 91%, reflecting rapid head adaptation. The stage boundary at epoch 6 produces a sharp improvement as full fine-tuning begins. Validation accuracy reaches its best value of 99.57% at epoch 12. Early stopping triggers at epoch 17. The close alignment between training and validation curves throughout Stage 2 confirms that label smoothing, dropout, MixUp/CutMix augmentation, and EMA effectively control overfitting despite fine-tuning 86M parameters.

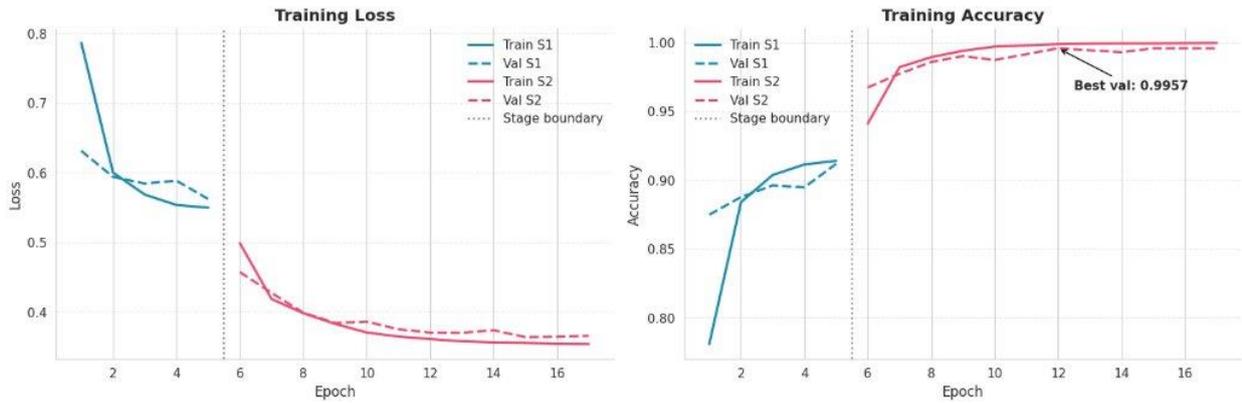

*Figure 4: Two-stage training history- Loss (left) and Accuracy (right). Best val accuracy: 99.57%*

## 5.2 Test Set Performance

Table 1 presents per-class results on the 703-image test set. Overall accuracy is 99.29%, with macro-average precision, recall, and F1-score all at 99.25–99.26%. The healthy class achieves perfect classification (all metrics = 1.0000), consistent with normal anatomy being the most morphologically distinct class. Meningioma achieves 100% recall with 97.06% precision 2 pituitary cases were misclassified as meningioma, an anatomically adjacent location. Glioma achieves 98.15% recall with perfect precision; 3 glioma cases were missed but all glioma predictions were correct.

*Table 1: Per-class classification performance on the test set (n = 703)*

| Class | Precision | Recall | F1-Score | Support |
|---|---|---|---|---|
| Glioma | 1.0000 | 0.9815 | 0.9907 | 162 |
| Healthy | 1.0000 | 1.0000 | 1.0000 | 200 |
| Meningioma | 0.9706 | 1.0000 | 0.9851 | 165 |
| Pituitary | 1.0000 | 0.9886 | 0.9943 | 176 |
| **Macro Avg** | **0.9926** | **0.9925** | **0.9925** | **703** |
| **Weighted Avg** | **0.9931** | **0.9929** | **0.9929** | **703** |
| **Overall Acc** | | **0.9929** | | **703** |

## 5.3 Confusion Matrix Analysis

A confusion matrix is a contingency table that captures how often the actual (true) class labels align with the model's predicted labels, summarizing the agreement and discrepancies between them (Mbonu et al, 2025). The confusion matrices in Figure 5 show 159/162 glioma, 200/200 healthy, 165/165 meningioma, and 174/176 pituitary correctly classified only 5 misclassifications out of 703 samples. 3 glioma and 2 pituitary samples are classified as meningioma. The row-normalized percentage matrix confirms no class falls below 98.1% recall. The absence of any glioma-healthy or pituitary-healthy confusions is particularly significant, as these represent the most clinically dangerous failure modes in a screening application.

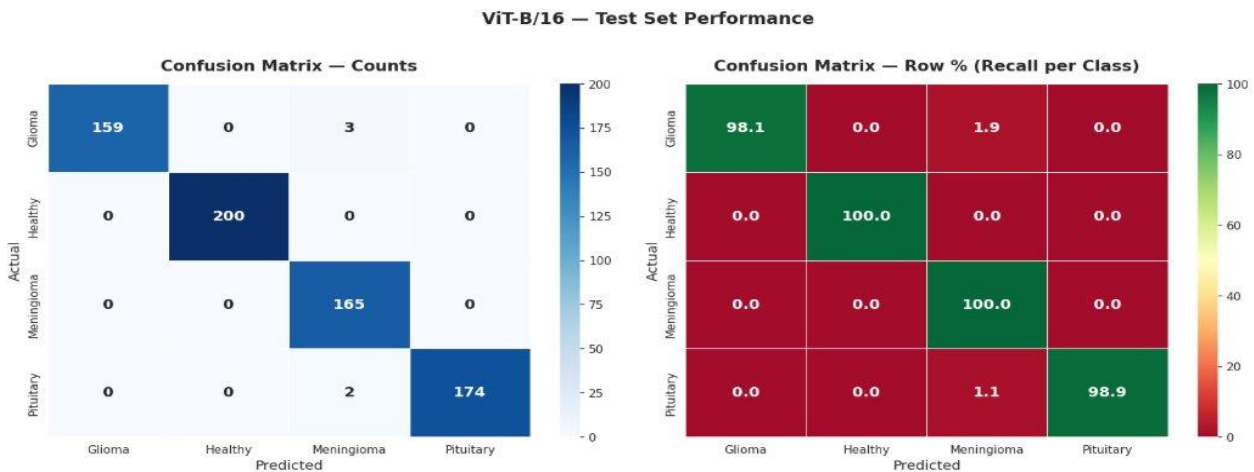

*Figure 5: Confusion Matrix — Raw counts (left) and row-normalized recall percentages (right)*

## 5.4 Attention Rollout Visualization

Figure 6 presents Attention Rollout heatmaps for representative test samples. For glioma cases, attention concentrates on diffuse hypointense or hyperintense infiltrative regions in the frontal and temporal lobes, with secondary attention on the tumor-brain interface. For healthy cases, attention distributes across symmetric deep brain structures including the corpus callosum and basal ganglia, reflecting the model's focus on bilateral symmetry as a feature of normal anatomy. For meningioma, attention strongly localizes the well-circumscribed peripheral mass with dural tail correctly identifying the extraaxial tumor location. The clinical coherence of these heatmaps establishes Attention Rollout as a practical interpretability tool for neuroimaging AI.

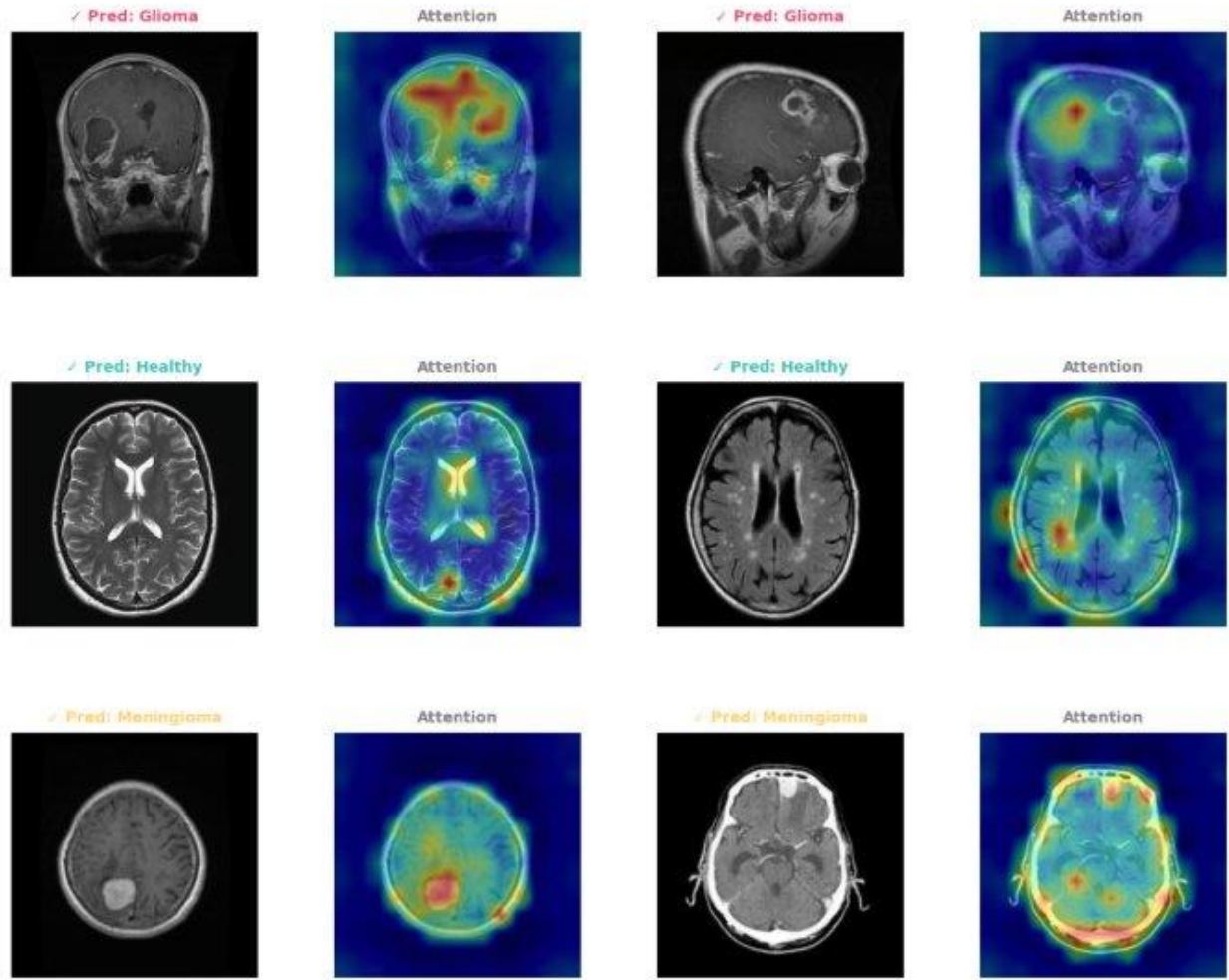

*Figure 6: Attention Rollout — original MRI (left) and attention heatmap overlay (right) per class*

## 5.5 Comparison with Prior Work

Table 2 compares the proposed method against representative prior work. The ViT-B/16 pipeline achieves 99.29%, surpassing the strongest VIT baseline (VIT: 98.70%) by +1.19 percentage points. On a screening dataset of 7,023 scans, this improvement corresponds to approximately 120 fewer misclassifications per 7,000 patients screened. Furthermore, the proposed method is the only approach in the comparison providing native attention-based interpretability without post-hoc approximations.

*Table 2: Comparison with prior work on brain tumor MRI classification*

| Study | Model | Dataset | Accuracy |
|---|---|---|---|
| Abiwinanda et al. (2019) | Custom CNN | 3,064 imgs | 84.19% |
| Sultan et al. (2019) | Deep CNN | 3,064 imgs | 96.13% |
| Deepak & Ameer (2019) | GoogleNet + SVM | 3,064 imgs | 97.10% |
| Sankari et al. (2025) | VIT | 7,023 imgs | 98.70% |
| **Proposed (ViT-B/16)** | **ViT-B/16 + CLAHE + MixUp/CutMix** | **7,023 imgs** | **99.29%** |

# 6. CONCLUSION

This paper presented a comprehensive deep learning framework for automated brain tumor classification from MRI scans using Vision Transformer (ViT-B/16), achieving test accuracy of 99.29% and macro F1-score of 99.25% across four classes on 7,023 MRI scans outperforming all surveyed CNN-based baselines. The key contributions enabling this performance are: CLAHE preprocessing enhancing tumor boundary contrast; MixUp and CutMix augmentation forcing class-discriminative feature learning; two-stage fine-tuning preserving pretrained feature quality while adapting to the MRI domain; EMA weight averaging stabilizing optimization; TTA providing inference-time accuracy improvement; and Attention Rollout producing clinically interpretable spatial heatmaps localizing the brain regions driving each prediction.

The Attention Rollout analysis confirms the model's spatial attention is clinically coherent: glioma cases focus on diffuse infiltrative regions, meningioma attends to well-circumscribed peripheral masses, and healthy cases distribute attention over symmetrical deep brain structures. This interpretability is a prerequisite for clinical deployment, where clinicians must audit and validate AI-generated diagnoses.

Future work will explore: (1) external validation on multi-center MRI datasets to assess generalizability across acquisition protocols; (2) integration of multi-sequence MRI modalities (T1, T2, FLAIR, DWI) as multi-channel inputs; (3) extension to tumor grading (WHO grade I–IV); (4) investigation of ViT-L/16 variants for the accuracy-efficiency tradeoff; and (5) prospective clinical validation with neuroradiologists to assess utility as a second-reader diagnostic tool.